\pdfoutput=1
\documentclass[twoside,11pt]{article}
\usepackage[preprint]{dmlr2e}

\usepackage{microtype}
\usepackage{url}
\usepackage{booktabs}
\usepackage{amsmath}
\usepackage{multirow}
\usepackage{xcolor}
\usepackage{subcaption}
\usepackage{enumitem}
\usepackage{cleveref}
\usepackage{tikz}
\usetikzlibrary{shapes,arrows.meta,positioning,fit,backgrounds,calc}
\usepackage{listings}
\usepackage{array}
\usepackage{algorithm}
\usepackage{algpseudocode}
\graphicspath{{figures/}}

\definecolor{einblue}{HTML}{0072B2}
\definecolor{einorange}{HTML}{E69F00}
\definecolor{eingreen}{HTML}{009E73}
\definecolor{einred}{HTML}{D55E00}
\definecolor{einpurple}{HTML}{CC79A7}
\definecolor{eingray}{HTML}{999999}
\definecolor{lightgray}{HTML}{F5F5F5}
\definecolor{darkblue}{rgb}{0, 0, 0.5}

\lstdefinelanguage{Ein}{
  keywords={trace, sum, max, min, diag, transpose, ones, zeros, eye, reach, desc, tc, has_path, select, ge, gt, eq},
  keywordstyle=\color{einblue}\bfseries,
  comment=[l]{\#},
  commentstyle=\color{eingray}\itshape,
  stringstyle=\color{eingreen},
  basicstyle=\ttfamily\scriptsize,
  breaklines=true,
  frame=single,
  backgroundcolor=\color{lightgray},
}

\title{ClosureBench: A Constructive Benchmark \\
for Compositional Graph Reasoning}

\author{\name Stefano Goria \email stefano@aimresearchlab.com \\
       \addr AIM Research Lab\\
       \url{aimresearchlab.com}}

\editor{}

\ShortHeadings{ClosureBench: Compositional Graph Reasoning}{Goria}
\firstpageno{1}
\def\openreview{}

\makeatletter
\def\@maketitle{\vbox{\hsize\textwidth
 \linewidth\hsize \vskip \beforetitskip
 {\begin{center} \Large\bf \@title \par \end{center}} \vskip \aftertitskip
 {\def\and{\unskip\enspace{\rm and}\enspace}%
  \def\addr{\small\it}%
  \def\email{\hfill\small\sc}%
  \def\name{\normalsize\bf}%
  \def\AND{\@endauthor\rm\hss \vskip \interauthorskip \@startauthor}%
  \@startauthor \@author \@endauthor}
  \vskip \aftermaketitskip
}}
\makeatother

\begin{document}

\maketitle

% ============================================================
% ABSTRACT
% ============================================================
\begin{abstract}
Large language models fail on multi-step compositional reasoning, but measuring that failure is hard, because new models are trained on the benchmarks used to evaluate them. A fixed test set becomes a memorisation check soon after release. \emph{Constructive} benchmarks avoid this by generating instances on demand. We introduce \textsc{ClosureBench}, a constructive benchmark for graph-relational logical reasoning. Each task is built from explicit primitives (reachability, degree, set operations, connectivity, aggregation), and its reference answer is computed by executing code that implements that logic exactly. Ground truth is therefore verified, and the supply of fresh instances is unlimited. The benchmark spans 26 task categories at three compositional levels, with three independent difficulty axes: graph size, edge density, and query depth.

We evaluate models from 1.5B open weights to frontier systems (o3, GPT-4.1, Gemini~2.5, Claude Sonnet~4). Accuracy falls as graph size and query depth increase, and the two axes interact. The difficulty does not lie in the surface form, since it persists when the graph is given as a JSON edge list or an adjacency matrix rather than prose, nor in the reasoning rule, which models state correctly. It lies in carrying that rule out over the graph across many steps. A 4B model fine-tuned to emit verified programs instead of answers stays nearly flat across compositional levels, while every frontier model degrades. o3 falls from 96\% on atomic queries to 82\% on the most compositional; the 4B model holds at 93\% at a fraction of the token cost. The program offloads multi-step execution to a runtime, and the model's remaining errors are almost entirely misread edges. Constructive generation also supports a direct memorisation check, comparing accuracy on seen and fresh instances.
\end{abstract}

\begin{keywords}
  graph reasoning, compositional generalisation, benchmarks, data contamination, program synthesis
\end{keywords}

% ============================================================
% 1. INTRODUCTION
% ============================================================
\section{Introduction}
\label{sec:intro}

Consider a compliance analyst at a bank who must answer a routine but consequential question: can money move from a given customer to a sanctioned entity through any chain of intermediaries? The transactions form a graph in which accounts are nodes and a transfer is a directed edge, and the question asks whether a directed path connects the customer to the sanctioned account. Answering it means chaining several steps: following transfers, tracking which accounts have been reached, and combining the partial results. A missed path is an undetected sanctions exposure; a spurious one freezes a legitimate account. This is the kind of multi-step, relational question that organisations increasingly pose to language models in natural language, and the kind on which language models are reported to be unreliable.

The evidence is consistent across settings. Models handle single-step inference well but degrade as a task composes more steps. A formal analysis of chain-of-thought finds accuracy falling as deductions chain~\citep{saparov2023prontoqa}; multi-hop question answering stays hard when questions genuinely require composing sub-questions rather than exploiting shortcuts~\citep{trivedi2022musique,ho2020wikimultihop}; dedicated logical-reasoning suites expose systematic errors on nested and first-order structure~\citep{hu2021folio,liu2020logiqa,yu2020reclor}; and grade-school arithmetic loses accuracy under superficial edits, a sign of pattern-matching rather than a robust procedure~\citep{gsmsymbolic2024}. The newest reasoning-tuned models narrow the gap without closing it. OpenAI's o1 improves on planning benchmarks but does not solve them~\citep{valmeekam2024planbench}, and on controlled puzzles the strongest reasoning models collapse once compositional complexity crosses a threshold, curtailing their own reasoning effort as problems grow harder~\citep{shojaee2025illusion}. Whether such failures reflect a genuine limit or the way we test for it is itself debated~\citep{benchpress2024}, which is a reason to measure them more precisely.

Precise measurement runs into two obstacles. First, benchmarks decay: once a test set is public it is absorbed into the next round of training, so a fixed set of questions soon measures recall rather than reasoning. GSM8K~\citep{cobbe2021gsm8k} and MMLU~\citep{hendrycks2021mmlu} have saturated, their scores can be inflated by training on leaked data~\citep{zeng2024decontamination}, and small surface edits already move them~\citep{gsmsymbolic2024}. Procedurally generated benchmarks address this by instantiating fresh instances on demand, as in the DyVal family~\citep{zhu2024dyval,zhu2024dyval2} and monthly-refreshed suites such as LiveBench~\citep{livebench2024}. Second, the benchmarks that do target graph reasoning (GraphQA~\citep{fatemi2023graphqa}, NLGraph~\citep{wang2023nlgraph}, GraphWiz~\citep{chen2024graphwiz}, GraphArena~\citep{tang2024grapharena}, GraCoRe~\citep{yuan2024gracore}, GraphInstruct~\citep{luo2024graphinstruct}, and the kinship-focused CLUTRR~\citep{sinha2019clutrr}) score answers by string matching~\citep{beyondbench2024} and vary difficulty along a single coarse axis, so they cannot isolate \emph{which} operation a model fails on or \emph{how much} composition breaks it; the choice of graph encoding alone can move accuracy by tens of points~\citep{fatemi2023graphqa}, confounding reasoning with input format.

The sanctions query is one instance of a general operation: the \emph{transitive closure} of a relation, the set of all pairs joined by a directed path. Closure recurs beyond finance. It decides whether a fault propagates through a dependency graph, whether a change reaches a downstream service, or who descends from whom in a lineage. Related operations recur just as widely: counting a node's connections (degree), intersecting the sets two sources reach, finding mutually reachable groups (strongly connected components), or aggregating a quantity over a reachable set. We call these graph-relational \emph{logic primitives}; each is a precise, checkable operation on a graph, and realistic questions compose a handful of them. Measuring reasoning this precisely means controlling \emph{which} primitives a question uses and \emph{how many} it composes, with a reference answer one can trust.

\textsc{ClosureBench}, named for that closure operation, provides both. Each task is a query over one or more logic primitives, and its reference answer is \emph{computed}, not curated. The primitive is executed as a short program of a few lines of tensor logic~\citep{domingos2025tensorlogic}, which expresses graph operations as contractions over the adjacency matrix, run in the Ein language\footnote{Ein is an open-source standalone language, implemented in Rust: \url{https://github.com/egolabs-ai/ein-lang}.} or, equivalently, in Python with NetworkX.\footnote{NetworkX, a Python library for graph algorithms: \url{https://networkx.org}.} Because the answer is computed it is exact, and the generator can produce unlimited fresh, contamination-free instances from independent structural and surface seeds. The benchmark spans 26 task categories at three compositional levels and exposes three independent difficulty axes (graph size, edge density, and query depth), so an evaluator can fix two and vary the third to localise where a model breaks. Models see only a natural-language question and return a JSON answer; they never see the program behind it.

We make three contributions:
\begin{enumerate}[leftmargin=*,itemsep=2pt,topsep=2pt]
\item \textsc{ClosureBench} itself: a constructive benchmark of 26 graph-relational categories built from explicit logic primitives, with programmatically computed ground truth, three independent difficulty axes, and surface control~(\S\ref{sec:design}). Because it supplies unlimited fresh instances, it also gives a direct check for memorisation, the gap between accuracy on seen and on freshly generated instances, which we use to separate recall from reasoning under fine-tuning~(\S\ref{sec:results:contamination}).
\item A diagnostic of where language models fail on these tasks: accuracy falls as graph size and query depth grow and the two interact, and the failure is in executing the computation over the graph rather than in reading the graph or stating the rule; it survives JSON and adjacency-matrix encodings and is not recovered by extra reasoning tokens~(\S\ref{sec:results:main}--\ref{sec:results:encoding}).
\item Program synthesis as a level-invariant, low-cost alternative. Emitting a program instead of an answer has driven recent gains in graph reasoning~\citep{zhu2025rethinking,wang2025g1}; here a 4B model fine-tuned to emit verified programs stays nearly flat across compositional levels, where every frontier model degrades, at a fraction of the token cost, with Ein and Python targets giving the same result~(\S\ref{sec:results:ein}).
\end{enumerate}

% ============================================================
% 3. CLOSUREBENCH DESIGN
% ============================================================
\section{ClosureBench Design}
\label{sec:design}

\textsc{ClosureBench} is built around four design principles, each addressing a failure mode of existing benchmarks. \textbf{(1)~Programmatic ground truth.} Every reference answer is computed by executing the task's logic primitives as a program, which eliminates human labelling errors. \textbf{(2)~Constructive generation.} New instances are generated on demand, so the benchmark never runs out of fresh evaluation data. \textbf{(3)~Independent complexity axes.} Three orthogonal parameters let evaluators isolate specific failure modes. \textbf{(4)~Domain grounding.} Tasks are framed in real-world verticals to test whether models handle contextualised graph descriptions.

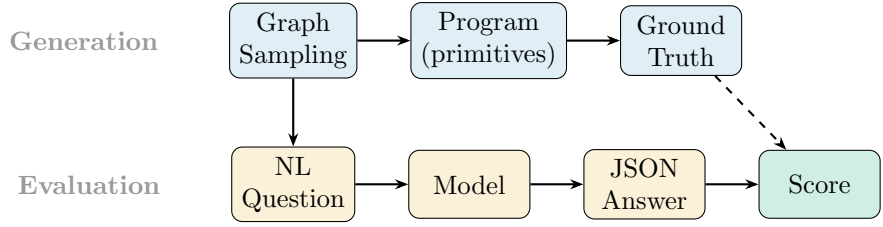
\begin{figure}[t]
\centering
\begin{tikzpicture}[
  node distance=0.6cm and 0.7cm,
  box/.style={draw, rounded corners, minimum height=0.9cm, minimum width=1.6cm, align=center, font=\small},
  phase/.style={font=\small\bfseries, text=eingray},
  arr/.style={-{Stealth[length=5pt]}, thick},
]
% Generation phase (top row)
\node[phase] (gen) {Generation};
\node[box, right=0.8cm of gen, fill=einblue!12] (sample) {Graph\\Sampling};
\node[box, right=of sample, fill=einblue!12] (einprog) {Program\\(primitives)};
\node[box, right=of einprog, fill=einblue!12] (gt) {Ground\\Truth};
\draw[arr] (sample) -- (einprog);
\draw[arr] (einprog) -- (gt);

% NL question branch from sampling
\node[box, below=0.9cm of sample, fill=einorange!15] (nlq) {NL\\Question};
\draw[arr] (sample) -- (nlq);

% Evaluation phase (bottom row)
\node[phase, left=0.8cm of nlq] (eval) {Evaluation};
\node[box, right=of nlq, fill=einorange!15] (model) {Model};
\node[box, right=of model, fill=einorange!15] (json) {JSON\\Answer};
\node[box, right=of json, fill=eingreen!18] (score) {Score};
\draw[arr] (nlq) -- (model);
\draw[arr] (model) -- (json);
\draw[arr] (json) -- (score);
\draw[arr, dashed] (gt) -- (score);
\end{tikzpicture}
\caption{The \textsc{ClosureBench} pipeline. \emph{Generation} (top): graphs are sampled and a program over the task's primitives computes verified ground truth (we use Ein, \S\ref{sec:design:ein}). \emph{Evaluation} (bottom): models receive only natural-language questions and produce JSON answers scored against precomputed ground truth. Models never see the programs.}
\label{fig:pipeline}
\end{figure}

\subsection{Reasoning Primitives and Query Types}
\label{sec:design:primitives}

Every \textsc{ClosureBench} task is built from a small set of graph-relational primitives over a graph's adjacency matrix. We define the primitives and the query types they induce below, and defer their computation to \S\ref{sec:design:ein}.

Represent a directed graph on $n$ nodes by its adjacency matrix $A \in \{0,1\}^{n\times n}$, with $A[i,j]{=}1$ iff there is an edge $i{\to}j$. Every reference answer is a function of $A$ and a few designated query nodes. The primitives are:

\begin{itemize}[leftmargin=*,itemsep=3pt,topsep=2pt]
\item \textbf{Degree.} The out-degree of node $i$ is the row sum $d_{\mathrm{out}}(i)=\sum_j A[i,j]$; the in-degree is the column sum. \emph{Query:} ``Alice pays Bob, Carol, and Dave. How many accounts does Alice pay?'' ($d_{\mathrm{out}}(\text{Alice}){=}3$).
\item \textbf{Reachability (transitive closure).} Node $i$ reaches $j$ if a directed path connects them. The closure matrix $\mathrm{TC}(A)$ has $\mathrm{TC}[i,j]{=}1$ iff $i$ reaches $j$; equivalently $\mathrm{TC}=\bigvee_{k=1}^{n-1} A^{k}$ under Boolean arithmetic. \emph{Query:} ``$X$ supplies $Y$, and $Y$ supplies $Z$; is there a supply path from $X$ to $Z$?''
\item \textbf{Reachable set.} $R(X)=\{j:\mathrm{TC}[X,j]{=}1\}$ is the set of nodes reachable from $X$. \emph{Query:} ``How many nodes are reachable from $X$?'' ($|R(X)|$).
\item \textbf{Set operations.} Queries combine reachable sets with intersection $R(X)\cap R(Y)$, difference $R(X)\setminus R(Y)$, or complement. \emph{Query:} ``Which nodes can $X$ reach that $Y$ cannot?'' ($R(X)\setminus R(Y)$).
\item \textbf{Triangles.} For an undirected graph the triangle count is $\mathrm{trace}(A^3)/6$ (each triangle closes six length-3 walks). \emph{Query:} ``How many groups of three nodes that all connect to one another are there?''
\item \textbf{Strongly connected components (SCC).} Nodes $i,j$ share an SCC iff each reaches the other, $\mathrm{TC}[i,j]\wedge \mathrm{TC}[j,i]$. \emph{Query:} ``Can $X$ and $Y$ each reach the other?''
\item \textbf{Ancestry and kinship.} On a directed acyclic parent$\to$child graph, ancestry is reachability, and kinship terms follow from it: an ancestor of $X$ reaches $X$; siblings share a parent; cousins share a grandparent. \emph{Query:} ``Eve is a parent of Carol and Tina; who is the oldest ancestor of Tina?'', or ``Are Carol and Dave cousins?''
\item \textbf{Aggregation.} A scalar summary (sum, mean, or maximum) over a set, such as the mean out-degree over $R(X)$, $\tfrac{1}{|R(X)|}\sum_{i\in R(X)} d_{\mathrm{out}}(i)$. \emph{Query:} ``Among the nodes reachable from $X$, what is their average number of outgoing links?''
\end{itemize}

A \emph{query type} is defined by which primitives it invokes and how they compose (\Cref{tab:primitives}). An \emph{atomic} query invokes one primitive: ``can $X$ reach $Y$?'' is a single closure lookup. A \emph{chain} feeds one primitive's output into the next: ``how many nodes can $X$ reach?'' is reachability followed by a count over $R(X)$. A \emph{composition} adds control flow: ``if $X$ can reach $Y$, report $|R(X)|$, otherwise $-1$'' runs a reachability test, branches on the result, then counts. The taxonomy (\S\ref{sec:design:taxonomy}) groups tasks by this structure, and the complexity axes (\S\ref{sec:design:complexity}) vary it independently.

\begin{table}[t]
\centering
\caption{Reasoning primitives, the questions they answer, and the categories that use them. $A$ is the adjacency matrix; $R(X)$ is the set of nodes reachable from $X$.}
\label{tab:primitives}
\small
\begin{tabular}{@{}p{2.1cm}p{3.1cm}p{3.4cm}p{3.1cm}@{}}
\toprule
\textbf{Primitive} & \textbf{Definition} & \textbf{Example question} & \textbf{Categories} \\
\midrule
Degree & $\sum_j A[i,j]$ & how many does $X$ point to? & degree\_count, degree\_max \\
Reachability & $\mathrm{TC}[X,Y]$ & can $X$ reach $Y$? & reachability, has\_path \\
Reachable-set size & $|R(X)|$ & how many can $X$ reach? & reach\_then\_count, reach\_then\_filter \\
Set operations & $R(X)\cap R(Y)$, $R(X)\setminus R(Y)$ & reachable from $X$ but not $Y$? & set\_intersect, set\_difference, negative\_reach \\
Triangles & $\mathrm{trace}(A^3)/6$ & how many triangles? & triangle\_count, triangle\_in\_subgraph \\
Connectivity (SCC) & $\mathrm{TC}[i,j]\wedge\mathrm{TC}[j,i]$ & are $X,Y$ mutually reachable? & scc\_same, scc\_then\_count \\
Ancestry/kinship & reachability on a DAG & who is an ancestor of $X$? & ancestor, sibling, cousin, kinship\_chain \\
Aggregation & $\mathrm{sum}/\mathrm{avg}/\max$ over a set & average degree of $R(X)$? & aggregate\_over\_set, mixed\_domain \\
Composition & branch/compose the above & if reachable, count; else $-1$ & conditional, multi\_query, path\_and\_compare \\
\bottomrule
\end{tabular}
\end{table}

\subsection{Task Taxonomy}
\label{sec:design:taxonomy}

\textsc{ClosureBench} organises 26 categories into three levels of compositional difficulty (\Cref{tab:taxonomy}); Appendix~\ref{app:categories} lists all of them with each category's defining operation.

\textbf{L1 (Atomic, 11 categories).} Each task applies a single primitive: triangle count, reachability, negative reachability, degree count, maximum degree, same-SCC membership, set intersection, set difference, and the kinship relations ancestor, sibling, and cousin. Each isolates one graph-relational operation.

\textbf{L2 (Chain, 8 categories).} Two or three primitives where one feeds the next: reach-then-count, reach-then-filter, kinship chain, triangles in a reachable subgraph, path-and-compare, SCC-then-count, degree-then-reach, and intersect-then-size. ``How many nodes can $X$ reach?'' is reachability followed by a count.

\textbf{L3 (Composition, 7 categories).} Three to five primitives with branching or multiple queries: multi-query, conditional, aggregate-over-set, complex kinship, graph comparison, chain-of-filters, and mixed-domain. ``If $X$ can reach $Y$, return the size of its reachable set, otherwise $-1$'' combines a reachability test, a branch, and a count.

\begin{table}[t]
\centering
\caption{Task taxonomy. Each level increases the number of chained reasoning operations. L1--L3 are included in the current release (1{,}300 instances total). SCC: strongly connected component.}
\label{tab:taxonomy}
\small
\begin{tabular}{@{}llccp{3.3cm}@{}}
\toprule
\textbf{Level} & \textbf{Name} & \textbf{Steps} & \textbf{Test} & \textbf{Representative Categories} \\
\midrule
L1 & Atomic & 1 & 550 & triangle count, reachability, degree, ancestor, set ops \\
L2 & Chain & 2--3 & 400 & reach-then-count, SCC-then-count, kinship chain \\
L3 & Composition & 3--5 & 350 & conditional, multi-query, aggregate-over-set \\
\bottomrule
\end{tabular}
\end{table}

\subsection{Three Independent Complexity Axes}
\label{sec:design:complexity}

Each \textsc{ClosureBench} instance is parameterised by three orthogonal complexity dimensions:

\textbf{Graph size ($n$):} the number of entities (nodes). Increasing $n$ makes the adjacency description longer and enlarges the space the model must track, without changing the computational structure. The held-out split uses $n\in\{4,\dots,10\}$; the size-scaling study (\S\ref{sec:results:main}) pushes $n$ to 20 on a dedicated sweep.

\textbf{Edge density ($\rho$):} the fraction of possible edges present ($\rho \in [0.1, 0.5]$). Sparse graphs ($\rho = 0.1$) have few connections and simple reachability; dense graphs ($\rho = 0.5$) produce more triangles, larger reachable sets, and more complex SCC structures.

\textbf{Query depth ($d$):} the number of chained reasoning operations, corresponding to levels L1--L3. Increasing $d$ adds compositional steps without changing the underlying graph.

These axes are orthogonal by construction: increasing $n$ does not change $d$; increasing $\rho$ does not change $n$. This enables controlled ablation studies that isolate specific failure modes, for instance testing whether a model that succeeds on 4-node graphs fails on 12-node graphs with the same query type, or whether density affects triangle counting but not reachability.

\subsection{Surface Variation}
\label{sec:design:surface}

Instance generation separates into \emph{structural} and \emph{surface} components controlled by independent seeds, following GSM-Symbolic's finding that surface perturbations degrade memorisation-dependent performance~\citep{gsmsymbolic2024}.

\textbf{Structural seed:} controls graph topology (which nodes connect to which) and therefore the reference answer. Two instances with the same structural seed but different surface seeds have identical answers but different natural-language realisations.

\textbf{Surface seed:} controls presentation features that do not affect the answer:
\begin{itemize}[itemsep=1pt,topsep=2pt]
\item \emph{Node labels} from disjoint pools: letters (A--Z), personal names (Alice--Vic), server identifiers (srv-00--srv-29), or animal names (ant--zebu).
\item \emph{Domain frames} (16 total): fraud detection, supply chain, cybersecurity, compliance, kinship, healthcare, logistics, social media, academia, banking, real estate, telecommunications, energy, government, transport, and abstract, each with domain-specific vocabulary (``transferred funds to'' vs.\ ``routes to'' vs.\ ``cites'').
\item \emph{Presentation style} (5 variants): flat, grouped by source, grouped by target with reversed verbs, randomly mixed, or varied per-source phrasing.
\item \emph{Edge ordering}: the order in which edges appear in the question text.
\end{itemize}

The combinatorial space is large. For a 10-node graph, each label pool provides $10^{10}$+ permutations; combined with 16 domains, 5 styles, and $m!$ edge orderings, a surface seed indexes into a space exceeding $10^{15}$ distinct realisations of the same graph structure. An evaluator can always generate a fresh surface configuration that shares no tokens with any training data, providing a structural defence against data contamination.

\subsection{Generating and Verifying Instances}
\label{sec:design:ein}

Generation runs offline (top of \Cref{fig:pipeline}). For each instance the generator samples a category and a structural triple $(n,\rho,d)$, draws a graph with those parameters, and computes the reference answer by executing the category's program over the graph; an independent surface seed (\S\ref{sec:design:surface}) then verbalises the graph and question as natural-language text. At evaluation time (bottom of \Cref{fig:pipeline}) a model receives only that text and returns a JSON answer, scored against the precomputed reference. The five-step procedure is detailed in Appendix~\ref{app:generation}.

The answer is \emph{computed}, not labelled, so it is correct by construction; we additionally cross-check every released instance against a second, independently written Python+NetworkX implementation of each category. Our primary engine is \emph{Ein}, a tensor-logic language. \citet{domingos2025tensorlogic} argues that logical rules and Einstein summation are the same operation; the primitives of \S\ref{sec:design:primitives} are naturally tensor contractions over the adjacency matrix, so they express directly and compactly in Ein. We use it for three properties: (1)~a parser and type-checker make execution deterministic and verifiable; (2)~programs are compact, typically 4--8 lines (${\sim}$200 tokens), which later makes Ein an efficient program-synthesis target (\S\ref{sec:results:ein}); and (3)~it provides the graph primitives (transitive closure, reachability, degree, selection) as builtins. Triangle counting is four lines:
\begin{lstlisting}[language=Ein]
A = edges([[0,1,1],[1,0,1],[1,1,0]], 3)
A2[i,k] = A[i,j] A[j,k]
A3[i,j] = A2[i,k] A[k,j]
Result = trace(A3) / 6.0
\end{lstlisting}
\emph{Models never interact with Ein}: the program runs only during generation, so any model can attempt \textsc{ClosureBench} without programming knowledge.

\textbf{Data splits.} We draw two disjoint splits from the generator, each stratified at 50 instances per category: a \emph{training} split and a \emph{held-out} split of 1{,}300 instances each. All models are evaluated on the held-out split; the fine-tuning experiments of \S\ref{sec:results:contamination} and \S\ref{sec:results:ein} train on the training split and report on the held-out split. Because the held-out split is verbalised from fresh surface seeds, it shares no surface tokens with the training split, so a model's seen-vs-held-out accuracy gap reflects memorisation rather than transfer (\S\ref{sec:results:contamination}).

% ============================================================
% 4. EXPERIMENTS
% ============================================================
\section{Experimental Setup}
\label{sec:experiments}

Our experiments use \textsc{ClosureBench} as a lens on how different classes of language model fail on graph-relational reasoning. We run three kinds of experiment: we evaluate open-weight and frontier models off the shelf, to map where accuracy breaks down (\S\ref{sec:results:main}--\ref{sec:results:errors}); a \emph{memorisation probe} fine-tunes a model on the benchmark's own instances, to test whether the constructive design can detect contamination (\S\ref{sec:results:contamination}); and a \emph{program-synthesis} experiment fine-tunes a model to emit an executable program instead of an answer, to test whether the failure can be removed (\S\ref{sec:results:ein}). This section gives the setup common to all three; Appendix~\ref{app:reproducibility} and the released code reproduce every number.

In every experiment a model is shown a natural-language graph question and must return a single JSON object with an \texttt{"answer"} field. For example, ``\emph{srv-0 routes to srv-2; srv-2 routes to srv-4; is there a route from srv-0 to srv-4?}'' expects \verb|{"answer": true}|. The schema is described generically, with no per-category examples, so a model must infer the expected type from the wording, as a human evaluator would (verbatim prompts in Appendix~\ref{app:prompts}, the per-category schema in Appendix~\ref{app:scoring}). Where the provider supports it we evaluate in two modes, \emph{direct} (answer immediately) and \emph{chain-of-thought} (CoT: reason step by step, then answer). We decode at temperature~0 throughout, so runs are deterministic and reproducible.

\subsection{Models}
We span three groups. \emph{Open-weight models} of increasing size run locally on a single low-cost device (one NVIDIA GB10, DGX~Spark): Qwen2.5-1.5B, Qwen3.5-2B, Qwen3-4B, and Qwen3.5-9B.\footnote{HuggingFace repositories (pinned commits): \texttt{Qwen/Qwen2.5-1.5B-Instruct} (\texttt{989aa79}), \texttt{Qwen/Qwen3-4B} (\texttt{1cfa9a7}), \texttt{Qwen/Qwen3.5-2B} (\texttt{15852e8}), and \texttt{Qwen/Qwen3.5-9B} (\texttt{c202236}). The Qwen3.5 models use the \texttt{Qwen3\_5ForConditionalGeneration} architecture (multimodal-capable); we run text-only inference.} \emph{Frontier models} are queried through their providers' APIs (March~2026, default settings): GPT-4.1, GPT-4.1-mini, o3, Claude Sonnet~4, and Gemini~2.5~Flash; o3 is available in direct mode only. \emph{Fine-tuned models} are Qwen3-4B adapted with LoRA~\citep{hu2022lora} for the memorisation and program-synthesis experiments; the configuration is in \S\ref{sec:exp:ft}.

\subsection{Evaluation Set and Difficulty}
Unless stated otherwise, all models are evaluated on the same held-out split (\S\ref{sec:design:ein}): 1{,}300 instances, 50 per category, drawn across the full difficulty range, with graph size $n\in\{4,\dots,10\}$, edge density $\rho$ drawn continuously from $[0.1,0.5]$, and query depth $d$ set by each category's level. The per-level accuracies we report therefore average over $n$ and $\rho$; \S\ref{sec:results:main} breaks accuracy out along each axis. All models are evaluated on the full 1{,}300-instance split.

\subsection{Scoring}
We report \emph{lenient} accuracy throughout: an answer is correct if it matches the reference, with two documented tolerances for the source-inclusion ambiguity present in some question phrasings ($\pm 1$ for the affected count categories, $\pm$one element for the affected set categories). \emph{Strict} scoring removes both tolerances. The two track each other closely and preserve the ordering of methods; the gap is $0.0$--$0.3$pp for program-synthesis models, $0.9$--$1.6$pp for frontier CoT, and $4.5$--$6.3$pp for direct prompting and small open models. The scoring algorithm and every configuration under both conventions are in Appendix~\ref{app:scoring} (\Cref{tab:strict_vs_lenient}).

\subsection{Fine-tuning Experiments}
\label{sec:exp:ft}
Two experiments fine-tune a model. The \emph{memorisation probe} (\S\ref{sec:results:contamination}) fine-tunes the base model on the training instances with the reference \emph{answers} as targets: if this lifts accuracy on seen instances far above accuracy on fresh held-out instances, the gap measures how much the model has memorised rather than learned. The \emph{program-synthesis} experiment (\S\ref{sec:results:ein}) fine-tunes the model to emit an executable \emph{program} (in Ein or Python+NetworkX) whose execution yields the answer, testing whether offloading the computation removes the compositional failure of \S\ref{sec:results:main}.

Both use the same configuration: Qwen3-4B with LoRA ($r{=}16$, $\alpha{=}32$, dropout~0.05), learning rate $2\times10^{-4}$, batch size~4, 3 epochs, trained on the 1{,}300-instance training split and evaluated on the disjoint held-out split, on the same GB10 device. The memorisation probe is repeated over 3 seeds ($\{42,137,256\}$) and averaged. For program synthesis the model has no prior exposure to Ein; it learns the syntax from a 600-token specification appended to the system prompt. Ground truth is computed with the Ein runtime v0.3.0.

% ============================================================
% 5. RESULTS
% ============================================================
\section{Results}
\label{sec:results}

We report results in three parts, one per contribution: the memorisation measurement (\S\ref{sec:results:contamination}), the diagnostic of where models fail (\S\ref{sec:results:main}--\ref{sec:results:errors}), and program synthesis as an alternative (\S\ref{sec:results:ein}).

\subsection{Measuring Memorisation}
\label{sec:results:contamination}

\begin{figure}[t]
\centering
\includegraphics[width=0.55\linewidth]{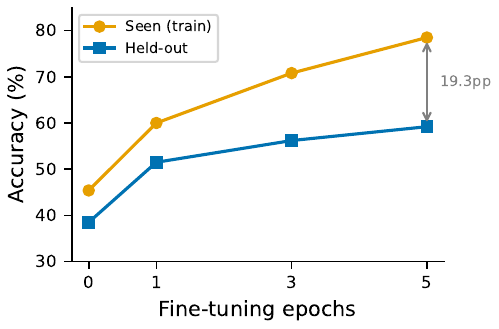}
\caption{Measuring memorisation (Qwen3-4B fine-tuned on 1{,}300 answer pairs). The gap between seen (78.5\%) and held-out (59.2\%) accuracy (19.3pp by epoch~5) is a memorisation signal that a fixed benchmark cannot measure.}
\label{fig:contamination}
\end{figure}

A fixed benchmark cannot separate a model that solved its questions from one that memorised them, since those questions may already be in the model's training data; a constructive benchmark can, because it supplies fresh instances. We fine-tune the base model on the 1{,}300 training instances with answer supervision (\S\ref{sec:exp:ft}) and track accuracy, over training epochs, on both the seen training instances and a disjoint held-out set. \Cref{fig:contamination} plots the two curves. Seen accuracy climbs to 78.5\% while held-out plateaus at ${\sim}$59\% after epoch~5, a 19.3pp gap. Held-out accuracy does rise from the base model's ${\sim}$34\% to 59.2\%, so fine-tuning yields some transferable gain; the 19.3pp gap is the part that does not transfer, and it widens with more training. The gap is for one model and configuration and its size will vary, so we treat the magnitude as illustrative, not a benchmark constant.

Because \textsc{ClosureBench} can generate unlimited fresh instances, evaluators can always produce a held-out set that the model has never seen. The seen-vs-held-out gap then serves as a \emph{memorisation signal}. A large gap (here 19.3pp) shows that part of the seen-set accuracy is memorisation that does not carry over to fresh instances. A fixed benchmark, where all instances are potentially in the training data, cannot make this measurement. Two caveats: this experiment tests supervised fine-tuning on aligned pairs and does not address subtler pretraining contamination, and the signal depends on the structural novelty of held-out instances.

\subsection{Where Do Models Fail?}
\label{sec:results:main}

\begin{figure}[t]
\centering
\includegraphics[width=\linewidth]{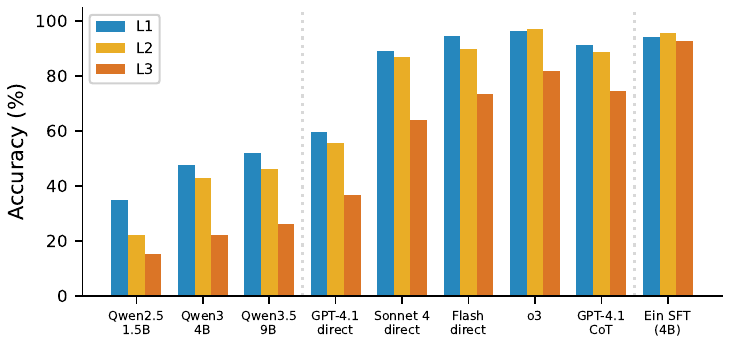}
\caption{Accuracy by compositional level across all evaluated models. Every model degrades from L1 to L3 except Ein~SFT (94\%$\to$93\%). Dotted lines separate open models, frontier API, and program synthesis.}
\label{fig:accuracy_by_level}
\end{figure}

\Cref{tab:main} presents accuracy across all models and levels. Every model degrades from L1 to L3, which establishes compositional depth as a difficulty axis. The L1--L3 drop (the ``compositional slope'') is 20--26pp for open models and 17--23pp for frontier models including o3; for Ein~SFT it is \textbf{1.3pp}. Program synthesis removes the compositional penalty that the natural-language (NL) reasoning approaches all carry. \Cref{fig:accuracy_by_level} plots the contrast.

\begin{table}[t]
\centering
\caption{Main results on \textsc{ClosureBench} (lenient scoring). All models are evaluated on the same 1{,}300-instance held-out split, spanning graph sizes $n{=}4$--$10$; accuracy is averaged over that range (\Cref{tab:size} breaks it out by $n$). All methods degrade from L1 to L3 except Ein~SFT, which stays nearly flat.}
\label{tab:main}
\small
\begin{tabular}{@{}lcccc@{}}
\toprule
\textbf{Model} & \textbf{L1} & \textbf{L2} & \textbf{L3} & \textbf{Overall} \\
\midrule
\multicolumn{5}{@{}l}{\textit{Open models (direct mode)}} \\
Qwen2.5-1.5B & 35.1 & 22.2 & 15.4 & 25.8 \\
Qwen3.5-2B & 40.2 & 31.8 & 23.1 & 33.0 \\
Qwen3-4B & 47.8 & 43.0 & 22.3 & 39.5 \\
Qwen3.5-9B & 52.0 & 46.2 & 26.3 & 43.3 \\
\midrule
\multicolumn{5}{@{}l}{\textit{Frontier API (direct mode)}} \\
GPT-4.1-mini & 58.2 & 49.5 & 34.9 & 49.2 \\
GPT-4.1 & 59.8 & 55.8 & 36.9 & 52.4 \\
Gemini 2.5 Flash & 94.5 & 89.8 & 73.4 & 87.4 \\
Claude Sonnet 4 & 89.1 & 87.0 & 64.0 & 81.7 \\
o3 & \textbf{96.4} & \textbf{97.2} & 82.0 & 92.8 \\
\midrule
\multicolumn{5}{@{}l}{\textit{Frontier API (chain-of-thought)}} \\
GPT-4.1-mini CoT & 91.1 & 88.0 & 74.9 & 85.8 \\
GPT-4.1 CoT & 91.5 & 89.0 & 74.6 & 86.2 \\
Claude Sonnet 4 CoT & 90.5 & 90.2 & 65.1 & 83.6 \\
Gemini 2.5 Flash CoT\textsuperscript{$\ddagger$} & 91.8 & 82.8 & 63.1 & 81.3 \\
\midrule
\multicolumn{5}{@{}l}{\textit{Program synthesis}} \\
\textbf{Ein SFT (Qwen3-4B)} & 94.2 & 95.8 & \textbf{92.9} & \textbf{94.3} \\
\bottomrule
\multicolumn{5}{@{}l}{\footnotesize\textsuperscript{$\ddagger$}CoT \emph{decreases} accuracy vs.\ direct mode (87.4\%); see \S\ref{sec:results:cost}.}
\end{tabular}
\end{table}

% Complexity scaling analysis (part of RQ2)
\label{sec:results:scaling}

\begin{figure}[t]
\centering
\includegraphics[width=\linewidth]{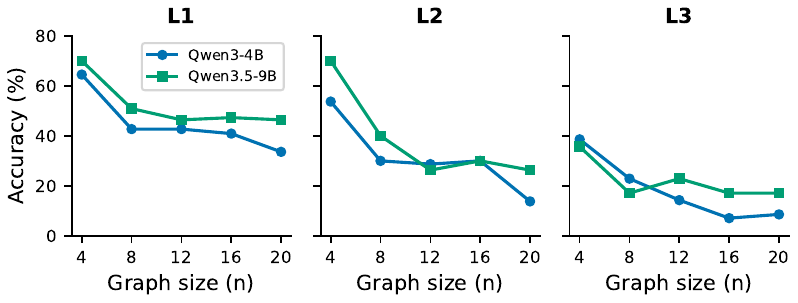}
\caption{Accuracy vs.\ graph size ($n$) at each compositional level, for the open models on a dedicated size sweep ($n$ up to 20). L1 degrades gently; L3 steeply. Size and depth compound rather than add.}
\label{fig:complexity_scaling}
\end{figure}

Beyond the level averages, accuracy depends on graph size, and the two interact. \Cref{fig:complexity_scaling} plots this for the open models on a dedicated size sweep ($n$ up to 20): L1 tasks degrade gently, while L3 tasks degrade steeply, so graph size and compositional depth compound rather than add. A drop of about 20pp appears between $n{=}4$ and $n{=}8$ for both Qwen3-4B and Qwen3.5-9B, after which accuracy declines gradually. The drop sits at nearly the same place for both model sizes, consistent with a limit in processing graph descriptions rather than a capacity limit, though we test only two small open models here; the L1--L3 gap widens from ${\sim}$26pp at $n{=}4$ to ${\sim}$30pp at $n{=}16$, where L3 falls to single digits.

The frontier models show the same size dependence on the held-out split, which spans $n\in\{4,\dots,10\}$ (\Cref{tab:size}). Overall accuracy falls with $n$ for every frontier model except o3, whose L1 and L2 stay near-perfect. The decline is driven by L3, and it takes the form of a threshold rather than a gradient: on small graphs ($n\le5$) frontier models answer L3 at 85--95\%, but past $n\approx5$ their L3 accuracy drops into a 55--70\% band (o3 higher, ${\sim}$82\%) and then plateaus rather than falling further. Edge density has a much weaker, non-monotonic effect (a few points from sparse to dense), so no single geometric axis orders the difficulty cleanly. Size-dependent degradation on compositional queries is thus a property of every NL-reasoning model we evaluate, open and frontier.

\begin{table}[t]
\centering
\caption{Held-out \emph{overall} accuracy (\%) by graph size $n$ (lenient). Every frontier model degrades with $n$ except o3, whose L1/L2 stay near-perfect. The decline is driven by the compositional (L3) tasks, which drop sharply once $n$ exceeds ${\sim}5$ and then plateau rather than declining smoothly; per-$n$ L3 estimates (${\sim}$50 instances each) are too noisy to tabulate. Modes: o3 and Gemini direct, GPT-4.1 and Claude CoT.}
\label{tab:size}
\small
\begin{tabular}{@{}lcccc@{}}
\toprule
\textbf{Model} & $n{=}4$ & $n{=}6$ & $n{=}8$ & $n{=}10$ \\
\midrule
o3 & 95 & 90 & 94 & 94 \\
Gemini 2.5 Flash & 92 & 89 & 85 & 79 \\
GPT-4.1 (CoT) & 92 & 88 & 85 & 75 \\
Claude Sonnet 4 (CoT) & 92 & 81 & 78 & 78 \\
\bottomrule
\end{tabular}
\end{table}

\subsection{Cost of Chain-of-Thought}
\label{sec:results:cost}

The frontier models reach their best accuracy only with chain-of-thought: it recovers much of the compositional loss above (GPT-4.1 rises from 52.4\% to 86.2\% overall). The reasoning tokens are not free, and whether that accuracy is affordable depends on how many are spent per answer, which \Cref{tab:token_cost} quantifies.

\begin{table}[t]
\centering
\caption{Token cost per answer and cost-efficiency (tokens per correct answer). CoT reasoning consumes 10--50$\times$ more tokens than direct mode, with superlinear scaling at higher compositional levels.\textsuperscript{$\star$}}
\label{tab:token_cost}
\small
\begin{tabular}{@{}llrrr@{}}
\toprule
\textbf{Model} & \textbf{Mode} & \textbf{Tok/ans} & \textbf{Acc.\ (\%)} & \textbf{Tok/correct} \\
\midrule
Qwen3-4B & Direct & 38 & 39.5 & 96 \\
Qwen3.5-9B & Direct & 52 & 43.3 & 120 \\
GPT-4.1 & Direct & 7 & 52.4 & 13 \\
Gemini Flash & Direct & 1{,}261 & 87.4 & 1{,}443 \\
GPT-4.1 & CoT & 431 & 86.2 & 500 \\
o3 & Direct & 844 & 92.8 & 910 \\
\midrule
\textbf{Ein SFT (4B)} & \textbf{Program} & \textbf{${\sim}$200} & \textbf{94.3} & \textbf{212} \\
\bottomrule
\end{tabular}

\vspace{2pt}
{\scriptsize All accuracies are overall (1{,}300 instances, lenient scoring).}
\end{table}

\begin{figure}[t]
\centering
\includegraphics[width=0.65\linewidth]{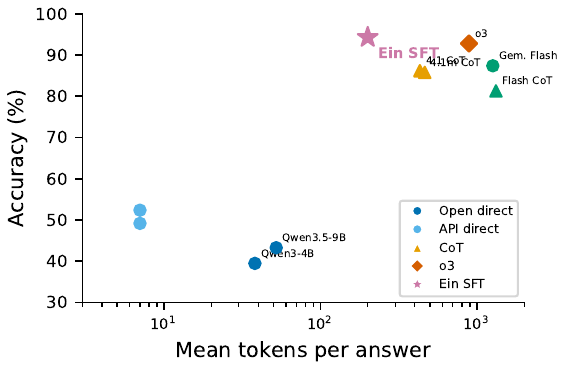}
\caption{Token efficiency: accuracy vs.\ mean tokens per answer (log scale). Ein SFT (star) achieves the highest accuracy at moderate token cost.}
\label{fig:token_cost}
\end{figure}

\Cref{fig:token_cost} visualises this trade-off: Ein SFT achieves the best accuracy--cost Pareto point, while CoT models cluster in the high-token, moderate-accuracy region.

Costs scale superlinearly with compositional depth: models generate more tokens per reasoning step at higher levels, compounded by error-recovery overhead (``Wait, let me reconsider\ldots'' segments that appear in 30--40\% of L3 CoT traces). As complexity grows, models spend more tokens per answer and get lower accuracy per token. GPT-4.1 CoT uses ${\sim}$431 tokens per answer, a 60$\times$ increase over direct mode, while only doubling accuracy. The Ein program-synthesis approach (\S\ref{sec:results:ein}) uses ${\sim}$200 tokens per answer at 94.3\% accuracy, the best cost-efficiency of any method evaluated.

CoT is not uniformly beneficial. For GPT-4.1, it adds +70pp on \texttt{aggregate\_over\_set} but costs $-$2pp on \texttt{kinship\_complex}, the category with the deepest traversal. For Gemini~2.5~Flash, CoT \emph{decreases} accuracy on 15 of 26 categories (up to $-$30pp on \texttt{reach\_then\_count}) and lowers overall accuracy from 87.4\% to 81.3\%. Two effects contribute. The model already reasons internally in direct mode (${\sim}$1{,}000 thinking tokens), so added thinking sometimes revises a correct answer; and under CoT it returns no parseable answer on 9.5\% of instances (\Cref{tab:format_error}), its reasoning trace exhausting the output budget before it emits JSON.

A rough compute comparison points the same way: Ein SFT generates ${\sim}$200 tokens from a 4B model, while frontier CoT generates 400--1{,}200 tokens from much larger models. We do not have parameter counts for the closed frontier models, so we do not quantify a combined tokens-times-parameters ratio; the token counts alone already favour the small program-synthesis model.

\subsection{Error Analysis}
\label{sec:results:errors}

The results so far show \emph{where} models fail; to see \emph{how}, we read the failing responses. Among the 89 L3 errors of GPT-4.1~CoT (our best frontier CoT model), three families account for 69\%; concrete examples are in Appendix~\ref{app:examples}.

\textbf{Direction confusion (24\%, multi\_query):} The model inconsistently parses directed edges. ``X reports to Y'' should encode as $X{\to}Y$ (out-degree of $X$), but the model sometimes counts edges pointing \emph{into} $X$, confusing out-degree with in-degree or total degree. For example, a node with out-degree~1 and in-degree~2 is reported as having degree~3. The reachability component is often correct; only the degree is wrong. The error is systematic across instances.

\textbf{Depth truncation (26\%, kinship\_complex):} The model finds the correct root ancestor but counts only 2--3 levels of descendants, systematically undercounting by 2--4. The distribution of errors is $\{-3{:}10, -4{:}6, -2{:}5\}$, consistent with a fixed depth limit. The model can trace a breadth-first traversal for a few steps but loses track of deeper branches.

\textbf{Accumulation drift (19\%, mixed\_domain):} Multi-step arithmetic (find SCC, collect degrees, compute average) accumulates small errors. Answers are off by 0.1--0.5 on floats. Each step is nearly correct, but errors compound across 4--5 operations.

The remaining 31\% are format mismatches (lists instead of dicts) and semantic ambiguities (source inclusion). Across all L3 failures, graph size strongly predicts difficulty: 2\% failure rate at $n{=}4$ vs.\ 36\% at $n{=}10$.

The model usually states the reasoning rule correctly; the failures are in carrying it out over the graph. Depth truncation and accumulation drift are execution errors. The model tracks a traversal or a running total for a few steps, then loses it, and the errors grow with the number of operations it must hold in working memory. Direction confusion is a separate, perceptual error: misreading the direction of an edge. Program synthesis (\S\ref{sec:results:ein}) offloads the execution to a runtime, which removes the first kind and leaves only the second.

\subsection{Program Synthesis as an Alternative}
\label{sec:results:ein}

The execution bottleneck (\S\ref{sec:results:errors}) motivates a different approach: train models to emit \emph{verified programs} whose execution produces the answer, instead of carrying out the computation in context. When the program is syntactically correct and type-checks, the Ein runtime returns a deterministic result, which moves the multi-step execution from the model to the runtime.

We fine-tune Qwen3-4B with LoRA ($r{=}16$, $\alpha{=}32$) on 1{,}300 (question, Ein~program) pairs for 3 epochs. The model has no prior exposure to Ein; it learns the complete syntax from a 600-token prompt specification appended to the system message. Training data and evaluation data are fully disjoint (1{,}300 / 1{,}300 instances), with no shared instances between splits.

\begin{table}[t]
\centering
\caption{Direct answering vs.\ program synthesis on the same model (Qwen3-4B, matched LoRA configuration). Trained on the same data, fine-tuning to emit answers memorises (19.3pp seen/held-out gap), while fine-tuning to emit programs generalises (3.5--3.7pp gap). The fair within-model comparison is the fine-tuned-direct row against the program-synthesis rows.}
\label{tab:ein}
\small
\begin{tabular}{@{}lcrrr@{}}
\toprule
\textbf{Method} & \textbf{Exec.\ rate} & \textbf{Seen} & \textbf{Held-out} & \textbf{Gap} \\
\midrule
Qwen3-4B (base, direct NL) & 100\% & 34.5\% & 34.2\% & --- \\
Qwen3-4B (answer SFT, direct) & 100\% & 78.5\% & 59.2\% & 19.3pp \\
Qwen3-4B (Python+NetworkX SFT) & 97.8\% & 96.1\% & 92.6\% & 3.5pp \\
Qwen3-4B (Ein SFT) & 99.6\% & \textbf{98.0\%} & \textbf{94.3\%} & 3.7pp \\
\bottomrule
\end{tabular}
\end{table}

The Ein model executes on 99.6\% of instances (1{,}295/1{,}300) and reaches 94.3\% held-out (1{,}226/1{,}300), with a 3.7pp seen/held-out gap (\Cref{tab:ein}). The controlled comparison is within the same model: the fine-tuned-direct model from \S\ref{sec:results:contamination}, trained on the same data with the same LoRA configuration but supervised on answers, reaches 59.2\% held-out with a 19.3pp gap. Supervising on programs rather than answers adds 35pp of held-out accuracy and shrinks the generalisation gap by 5$\times$, which indicates the model learns to write programs rather than to store input--output pairs. Held-out accuracy also exceeds zero-shot o3 (92.8\%), but that sets a task-specialised fine-tune against a general model prompted zero-shot, so we do not read it as evidence of a stronger model. The results we rely on are the near level-invariance (\Cref{tab:main}) and the low token cost (\S\ref{sec:results:cost}).

All 26 categories exceed 82\% on held-out; 21 exceed 90\%. The five below-90\% categories (\texttt{triangle\_count}, \texttt{degree\_count}, \texttt{triangle\_in\_subgraph}, \texttt{aggregate\_over\_set}, \texttt{mixed\_domain}) involve either counting operations sensitive to single-edge errors or compound operations whose multi-step descriptions make edge extraction harder. Both are the graph-reading error that remains once execution is offloaded (\S\ref{sec:results:errors}).

Among the 74 held-out errors, 61 (82\%) are graph-parsing mistakes: the model writes syntactically valid Ein programs with correct logic but misreads one or more edges from the natural-language description. Only 8 errors (11\%) involve incorrect program logic, and 5 (7\%) are execution failures. The remaining bottleneck is therefore NL comprehension, not program synthesis.

The approach works with other target languages. We fine-tune the same Qwen3-4B under identical hyperparameters ($r{=}16$, $\alpha{=}32$, 3 epochs, temperature~0) to emit Python+NetworkX programs. The Python baseline reaches 92.6\% held-out accuracy at a 97.8\% execution rate, with a 3.5pp seen/held-out gap, close to Ein's 3.7pp. The two targets are within 1.7pp overall: Python wins or ties on 17 of 26 categories and leads on L3 by $+2.6$pp, while Ein leads on atomic L1 queries and is more compact (${\sim}$128 vs.\ ${\sim}$156 generated tokens per answer). A zero-shot Python probe (GPT-4.1, 52 instances) reaches 78.8\% at ${\sim}$219 tokens, which locates the gain in the fine-tuning. We use Ein for its tensor-logic verification guarantees and its compactness as an SFT target; the effect holds for any verified program-synthesis target that offloads the multi-step execution.

Ein programs average ${\sim}$200 tokens per answer compared to ${\sim}$844 for o3 and ${\sim}$431 for GPT-4.1~CoT (\Cref{tab:token_cost}). Combined with higher accuracy, the Ein approach uses ${\sim}$212 tokens per \emph{correct} answer, 6$\times$ fewer than o3 and 6--10$\times$ fewer than frontier CoT models.

\subsection{Robustness to Graph Encoding}
\label{sec:results:encoding}

The compositional penalty (\S\ref{sec:results:errors}) could be an artefact of the \emph{natural-language} presentation, since a cleaner, structured encoding might remove it. We test this directly. On a 200-instance L1/L2/L3-uniform subset (whose NL accuracy matches the full set within 0.5pp), we re-render each graph under three encodings while holding the query sentence fixed: (i)~the natural-language enumeration used throughout the paper, (ii)~a JSON object with explicit \texttt{nodes} and \texttt{edges} lists, and (iii)~a labelled $N{\times}N$ adjacency matrix. We evaluate four open models and two frontier models in direct mode.

\begin{table}[t]
\centering
\caption{Accuracy (\%) across three graph encodings on the same 200-instance subset (direct mode, lenient scoring). The compositional penalty survives every encoding; encoding \emph{preference} is capacity-dependent (note the adjacency-matrix column reverses sign between small and frontier models).}
\label{tab:encoding}
\small
\begin{tabular}{@{}lrrr@{}}
\toprule
\textbf{Model} & \textbf{NL} & \textbf{JSON edges} & \textbf{Adj.\ matrix} \\
\midrule
Qwen2.5-1.5B      & 26.5 & 26.5 & 24.5 \\
Qwen3.5-2B        & 28.5 & 29.5 & 19.5 \\
Qwen3-4B          & 40.0 & 38.0 & 24.5 \\
Qwen3.5-9B        & 47.5 & 51.0 & 30.0 \\
GPT-4.1           & 53.0 & 59.5 & 44.0 \\
Gemini 2.5 Flash  & 84.5 & 88.5 & \textbf{91.0} \\
\bottomrule
\end{tabular}
\end{table}

\Cref{tab:encoding} yields three findings. \textbf{(1)~The bottleneck is not NL phrasing.} A clean JSON edge list is statistically tied with NL for weaker models (Qwen2.5-1.5B: 26.5 vs.\ 26.5; Qwen3-4B: 38.0 vs.\ 40.0) and gives capable ones a moderate gain ($+3.5$ to $+6.5$pp, e.g.\ GPT-4.1 53.0$\to$59.5). Handing the model a perfectly structured edge list does not rescue it; the difficulty is in carrying out the computation over the graph, not the surface form. \textbf{(2)~Encoding preference is capacity-dependent.} The adjacency matrix is the \emph{worst} encoding for every open model ($-2$ to $-17.5$pp, worst for the mid-size models that cannot track an $N{\times}N$ grid positionally on graphs up to $n{=}20$), yet the \emph{best} encoding for the strongest model (Gemini 2.5 Flash, $+6.5$pp over NL). An unambiguous, position-addressable matrix is too dense for weak models to parse but the cleanest input for a model with enough capacity to index it, which may explain why prior work disagrees on the ``best'' graph encoding~\citep{fatemi2023graphqa}. \textbf{(3)~The compositional cliff persists across encodings.} The L1$\rightarrow$L3 drop survives all three representations for every model except Gemini under the adjacency matrix, where it flattens to L2~$=97.0$, L3~$=84.8$. Encoding choice reshuffles \emph{where} errors land without removing the compositional penalty, which keeps compositional depth as the main difficulty axis. One caveat on the matrix column: Gemini 2.5 Flash is a thinking model whose default mode spends internal reasoning tokens, so its strong adjacency-matrix result reflects that additional compute as well as the encoding itself; we read the column as capacity-dependent rather than as evidence that matrices are intrinsically easier. The three findings are unchanged under strict scoring.

% ============================================================
% 6. DISCUSSION
% ============================================================
\section{Discussion}
\label{sec:discussion}

\textsc{ClosureBench} was built to test whether large language models execute compositional graph logic or only appear to. Accuracy is high on atomic queries and falls as operations compose, and every model we tested, open and frontier, shows the same decline from L1 to L3 (o3 from 96\% to 82\%, the others further).

The failure is not a lack of task understanding but of \emph{executing} the computation over the graph. Three results place it there. It survives clean structured encodings, since neither a JSON edge list nor an adjacency matrix rescues it, so it is not an artefact of natural-language phrasing. It grows with graph size and depth together, as a working-memory limit would, for frontier models as well as open ones; o3's overall accuracy is least affected, its atomic and chained sub-tasks staying near-perfect, but even its hardest queries degrade with size. The individual errors are traversal and accumulation mistakes over a graph the model has otherwise read correctly.

Fine-tuning does not fix this by teaching the logic. Supervising the model on the benchmark's answers raises seen-set accuracy while held-out accuracy plateaus, a 19.3pp gap, so the model memorises graph--answer pairs rather than learning to compute. The constructive design exposes this. With fresh instances always available the gap is measurable, which a fixed benchmark cannot do.

The same 4B model succeeds once it no longer has to hold the computation in its head. Fine-tuned to emit a short program in Ein or Python+NetworkX and let a runtime execute it, it stays nearly flat across compositional levels (1.3pp from L1 to L3), matches or exceeds the frontier models on the hardest queries, and uses a fraction of the tokens. Its remaining errors are almost all misread edges rather than wrong logic, which is what is left once execution is offloaded.

\textbf{Limitations.} (1)~\textsc{ClosureBench} covers only graph-theoretic tasks with algorithmically computable ground truth. (2)~The memorisation probe fine-tunes on answers for a single model and configuration; it is a diagnostic, not immunity, and does not address subtler pretraining contamination. (3)~We lack 30B+ open models to bridge small open and frontier API. (4)~The program-synthesis result uses one 4B base model and two program targets; whether the near level-invariance holds for other models and larger scales is untested. (5)~Some question templates leave the source-inclusion convention implicit; we report strict and lenient scores side by side (\Cref{tab:strict_vs_lenient}) rather than resolving every template.

\textsc{ClosureBench}, including generator, evaluation harness, and all model responses, is released at \url{https://github.com/egolabs-ai/closurebench}; because instances are generated on demand, an evaluator is never limited to the splits we ship.

\impact{\textsc{ClosureBench} is an evaluation artifact; its intended effect is more reliable measurement of compositional reasoning and of memorisation. The benchmark is fully synthetic and contains no personal or sensitive data, so it carries no direct privacy or fairness risk. The main foreseeable misuse is training on the released instances to inflate reported scores; the memorisation measurement in \S\ref{sec:results:contamination} is designed to detect this. The program-synthesis results point toward cheaper verified reasoning, which we read as a positive efficiency outcome rather than a capability risk.}

\acks{The author declares no competing interests, and this work received no specific external funding.}

% ============================================================
% BIBLIOGRAPHY
% ============================================================
\bibliography{closurebench}

% ============================================================
% APPENDIX
% ============================================================
\appendix

\section{Full Category Descriptions}
\label{app:categories}

\textbf{L1 (11 categories):} triangle\_count ($\mathrm{trace}(A^3)/6$), reachability (transitive closure, $\mathrm{TC}(A)[s,t]$), negative\_reach (set complement), degree\_count/max, scc\_same, set\_intersect, set\_difference, ancestor, sibling, cousin.

\textbf{L2 (8 categories):} reach\_then\_count, reach\_then\_filter, kinship\_chain, triangle\_in\_subgraph, path\_and\_compare, scc\_then\_count, degree\_then\_reach, intersect\_then\_size.

\textbf{L3 (7 categories):} multi\_query, conditional, aggregate\_over\_set, kinship\_complex, graph\_comparison, chain\_of\_filters, mixed\_domain.

\section{Sample Questions}
\label{app:sample_questions}

Below we show one example question from each compositional level. Models receive only the natural-language text and must produce a JSON answer. The generator enforces that every entity named in a question is declared in the node list, and the examples below are reproduced directly from generator output.

\textbf{L1 (reachability):}
\begin{quote}
\small
There are 6 stations: deer, cat, wren, puma, quail, lynx. lynx supplies power to deer. puma and quail draw power from deer. cat draws power from wren. wren supplies power to puma. Is there a path from lynx to quail?
\end{quote}

\textbf{L2 (reach\_then\_count):}
\begin{quote}
\small
There are 7 suppliers: newt, eagle, kiwi, ibis, zebu, owl, viper. eagle receives components from ibis. owl receives components from newt and zebu. newt ships to eagle and kiwi. kiwi ships to ibis. viper ships to eagle, kiwi, and zebu. How many suppliers can viper reach?
\end{quote}

\textbf{L3 (conditional):}
\begin{quote}
\small
There are 5 departments: kiwi, robin, bear, zebu, newt. bear reports to robin. zebu oversees bear. kiwi reports to bear and newt. If there is a reporting path from kiwi to robin, how many departments can kiwi reach? Otherwise return $-1$.
\end{quote}

\section{Per-Category Results}
\label{app:percategory}

\Cref{tab:percategory} reports Ein SFT held-out accuracy for all 26 categories.

\begin{table}[t]
\centering
\caption{Ein SFT (Qwen3-4B + LoRA) held-out accuracy by category (50 instances per category). Categories below 90\% are shown in \textcolor{einred}{red}.} % experiment 015
\label{tab:percategory}
\small
\begin{tabular}{@{}llr@{}}
\toprule
\textbf{Level} & \textbf{Category} & \textbf{Held-out (\%)} \\
\midrule
L1 & triangle\_count & \textcolor{einred}{82} \\
L1 & reachability & 98 \\
L1 & negative\_reach & 96 \\
L1 & degree\_count & \textcolor{einred}{84} \\
L1 & degree\_max & 94 \\
L1 & scc\_same & 96 \\
L1 & set\_intersect & 100 \\
L1 & set\_difference & 98 \\
L1 & ancestor & 96 \\
L1 & sibling & 96 \\
L1 & cousin & 96 \\
\midrule
L2 & reach\_then\_count & 100 \\
L2 & reach\_then\_filter & 98 \\
L2 & kinship\_chain & 92 \\
L2 & triangle\_in\_subgraph & \textcolor{einred}{86} \\
L2 & path\_and\_compare & 96 \\
L2 & scc\_then\_count & 100 \\
L2 & degree\_then\_reach & 96 \\
L2 & intersect\_then\_size & 98 \\
\midrule
L3 & multi\_query & 90 \\
L3 & conditional & 100 \\
L3 & aggregate\_over\_set & \textcolor{einred}{88} \\
L3 & kinship\_complex & 94 \\
L3 & graph\_comparison & 96 \\
L3 & chain\_of\_filters & 98 \\
L3 & mixed\_domain & \textcolor{einred}{84} \\
\bottomrule
\end{tabular}
\end{table}

\section{Example Ein Program}
\label{app:ein_example}

The following annotated example shows how Ein computes the reference answer for a ``reach\_then\_count'' task (L2): given a directed graph, count the number of nodes reachable from a source node.

\begin{lstlisting}[language=Ein]
# Input: 5-node directed graph, source = node 2
# Adjacency matrix (1 = edge exists)
A = edges([[0,1,0,0,0],
           [0,0,1,0,0],
           [0,0,0,1,1],
           [0,0,0,0,0],
           [0,0,0,0,0]], 5)

# Transitive closure: TC[i,j] = 1 iff i can reach j
TC = tc(A)

# Select row for source node 2
Row[j] = select(TC, 2, j)

# Count reachable nodes (excluding source)
Result = sum(Row)
\end{lstlisting}

\textbf{Walkthrough.} Line~1 declares a $5{\times}5$ adjacency matrix. The edges $2{\to}3$, $2{\to}4$, $0{\to}1$, $1{\to}2$, and $2{\to}3$ are encoded as 1s. The \texttt{tc} builtin computes the transitive closure via repeated Boolean matrix multiplication. \texttt{select} extracts row~2 (source node), and \texttt{sum} counts the non-zero entries. The result (here, \texttt{3}) becomes the ground-truth answer.

Key properties: (1)~the program is compact (6 lines, ${\sim}$80 tokens); (2)~every operation is deterministic; (3)~the type system ensures dimensional consistency (matrix $\times$ matrix $\to$ matrix); (4)~the runtime guarantees termination. The model being evaluated never sees this program; it receives only the natural-language question ``How many nodes can node~2 reach in the following graph?'' and must produce \texttt{\{"answer": 3\}}.

\section{Ein Syntax Specification}
\label{app:ein_syntax}

Ein is a minimal tensor-logic language with the following core constructs:

\textbf{Declarations.} Tensors are declared with explicit dimensions:
\begin{lstlisting}[language=Ein]
A = edges([[0,1],[1,0]], 2)    # 2x2 adjacency matrix
v = vec([1,0,1], 3)            # length-3 vector
s = 42.0                       # scalar
\end{lstlisting}

\textbf{Einstein summation.} Named indices express contractions:
\begin{lstlisting}[language=Ein]
C[i,k] = A[i,j] B[j,k]        # matrix multiply
v[i] = A[i,j] ones(3)[j]       # row sums
s = A[i,j] B[i,j]              # Frobenius inner product
\end{lstlisting}
Repeated indices on the right-hand side are summed over (Einstein convention). Free indices on the left determine the output shape.

\textbf{Builtins.} Ein provides domain-specific operations:
\begin{itemize}[itemsep=1pt,topsep=2pt]
\item \texttt{tc(A)} --- transitive closure (Boolean matrix power)
\item \texttt{reach(A, i)} --- nodes reachable from $i$
\item \texttt{desc(A, i)} --- descendants of $i$ (alias for reach)
\item \texttt{has\_path(A, i, j)} --- Boolean reachability test
\item \texttt{trace(A)} --- matrix trace ($\sum_i A_{ii}$)
\item \texttt{sum(v)}, \texttt{max(v)}, \texttt{min(v)} --- vector reductions
\item \texttt{select(A, i, j)} --- index a matrix element
\item \texttt{ge(a,b)}, \texttt{gt(a,b)}, \texttt{eq(a,b)} --- comparison predicates
\item \texttt{diag(A)}, \texttt{transpose(A)} --- structural operations
\item \texttt{ones(n)}, \texttt{zeros(n)}, \texttt{eye(n)} --- constructors
\end{itemize}

\textbf{Semantics.} Every program is a sequence of assignments evaluated top-to-bottom. The final assignment's value is the program output. All operations are over non-negative reals; Boolean values are represented as $\{0, 1\}$. Programs are guaranteed to terminate (no loops, no recursion) and produce exactly one scalar, vector, or matrix value.

\section{Baseline Prompts}
\label{app:prompts}

\textbf{System prompt (direct mode):}
\begin{quote}
\small\ttfamily
You are a reasoning assistant. Read the graph description carefully, reason about the structure, and provide your answer as a JSON object with an \texttt{"answer"} field. Do not include any explanation outside the JSON.
\end{quote}

\textbf{System prompt (CoT mode):}
\begin{quote}
\small\ttfamily
You are a reasoning assistant. Read the graph description carefully. Before providing your answer, think step by step about the graph structure and the computation needed. Then provide your final answer as a JSON object with an \texttt{"answer"} field.
\end{quote}

\textbf{System prompt (Ein code-gen mode):}

The complete Ein syntax reference is appended to the system message. The model receives this specification at every inference call; it is the \emph{only} source of Ein knowledge beyond the supervised training examples.

\begin{lstlisting}[language=Ein,basicstyle=\ttfamily\tiny]
Ein is a minimal tensor-logic language for graph computation.
Programs operate on adjacency matrices and produce scalar or vector results.

SYNTAX:
- A = edges([[0,1],[1,2]], N)  -- adjacency matrix from edge list, N nodes
- B[i,k] = A[i,j] A[j,k]     -- Einstein summation (repeated indices summed)
- Result = trace(A)            -- trace (sum of diagonal)
- Result = sum(A) / sum(A, 1)  -- sum all / along axis
- Result = tc(A)               -- transitive closure (reachability matrix)
- Result = reach(A, i)         -- nodes reachable from i, INCLUDING i
- Result = desc(A, i)          -- descendants of i, EXCLUDING i
- Result = has_path(A, i, j)   -- 1.0 if path exists, else 0.0
- Result = select(V, i, d)     -- select element i along dim d
- Result = ge/gt/eq(A, x)      -- comparison predicates (return 0/1)
- U = ones(N) / zeros(N)       -- constructors
- X = A * B / A + B / A - B    -- elementwise ops
- :print Result                -- output (MUST be last line)

RULES:
- Start with comment: // category_name
- Define A = edges(...) for the adjacency matrix
- Compute answer using tensor operations
- End with :print Result
\end{lstlisting}

\section{Reproducibility}
\label{app:reproducibility}

The experimental setup (models, decoding, prompting, scoring, and the fine-tuning configuration) is given in \S\ref{sec:experiments}; all API models were evaluated in March~2026 with provider-default settings.

\textbf{Repository.} The release at \url{https://github.com/egolabs-ai/closurebench} includes:
\begin{itemize}[itemsep=1pt,topsep=2pt]
\item The 1{,}300-instance evaluation set and all model responses (raw JSONL with reasoning traces)
\item Instance generator: \texttt{python -m closurebench.generator --seed <N>} produces unlimited fresh instances with verified Ein ground truth
\item Evaluation harness with lenient scoring: \texttt{python scripts/run\_full\_benchmark.py}
\item Ein and Python SFT training scripts (fine-tuned adapter weights will be released on the Hugging Face Hub)
\item All figures and tables are reproducible from the included data and scripts
\end{itemize}
The constructive nature of the benchmark means evaluators are not limited to our 1{,}300 instances: generating a fresh set with a new seed takes under one minute and produces instances with machine-verified ground truth.

\section{Datasheet, Licensing, and Maintenance}
\label{app:datasheet}

We summarise the release documentation for \textsc{ClosureBench}, following the datasheets-for-datasets framework of \citet{gebru2021datasheets}.

\textbf{Motivation and intended use.} \textsc{ClosureBench} is intended for evaluating compositional graph-relational reasoning in language models and for measuring memorisation through the seen-vs-held-out gap. It is a diagnostic benchmark, not a training corpus: using the released instances as training data defeats the memorisation measurement, so we discourage it.

\textbf{Composition.} Each instance is a self-contained (question, reference answer) pair with metadata (category, level, result type, structural and surface seeds, and the Ein and Python programs that compute the answer). The release contains the training and held-out splits (1{,}300 instances each) and a further 2{,}500 instances provided for downstream fine-tuning. All instances are synthetic and contain no personal, sensitive, or human-subject data.

\textbf{Collection and generation.} Instances are generated programmatically (Appendix~\ref{app:generation}); there is no human annotation, so there is no annotator bias or labelling error. Ground truth is computed by program execution and cross-checked between two independent implementations.

\textbf{Licensing.} The dataset and generator are released under CC~BY~4.0; the evaluation and training code under the MIT license.

\textbf{Author statement.} The authors bear all responsibility in case of any violation of rights for the released artifacts and confirm the licenses stated above. All released instances are synthetic and machine-generated; the release contains no third-party, personal, or human-subject data.

\textbf{Hosting and maintenance.} The data, generator, and evaluation harness are hosted in a public repository under a versioned release, with an archived snapshot (assigned a DOI) accompanying the camera-ready; the fine-tuned adapter weights will be released on the Hugging Face Hub. Because every instance is regenerable from the released code and seeds, the artifact does not depend on continued hosting of any single file. Corrections and issues are tracked in the repository.

\textbf{Reproducibility.} We follow the ML reproducibility checklist; code, data, hyperparameters, compute environment, and random seeds are documented in Appendix~\ref{app:reproducibility}.

\textbf{Ethics and broader impact.} The benchmark is synthetic and contains no personal data. Its intended effect is more reliable evaluation of reasoning. The main foreseeable misuse is training on the released instances to inflate scores, which the memorisation measurement (\S\ref{sec:results:contamination}) is designed to expose.

\section{Output Schema and Scoring}
\label{app:scoring}

\textbf{JSON output schema.} Every model is asked to emit a single JSON object whose only required field is \texttt{answer}. The value type depends on the task's \texttt{result\_type}, declared at generation time and fixed across all instances of a category. The full mapping is in \Cref{tab:json_schema}. The schema is described generically in the system prompt, with no per-category type hints or worked examples, so the model must infer the expected type from the question wording, as a human evaluator would. The parser accepts standard JSON and additionally tolerates one common CoT artefact: a single \texttt{\{"answer": ...\}} object embedded in a longer string; anything else is recorded as a format error.

\begin{table}[h]
\centering
\caption{JSON output schema. Each instance carries a single \texttt{result\_type}; the model's \texttt{"answer"} field must conform to the corresponding JSON value. Empty sets and zero counts are valid answers.}
\label{tab:json_schema}
\footnotesize
\begin{tabular}{@{}llp{3.0cm}p{3.7cm}@{}}
\toprule
\textbf{\texttt{result\_type}} & \textbf{JSON value} & \textbf{Example} & \textbf{Categories} \\
\midrule
\texttt{boolean} & \texttt{true} / \texttt{false} & \texttt{\{"answer": true\}} & reachability, scc\_same, conditional \\
\texttt{integer} & non-negative integer & \texttt{\{"answer": 3\}} & triangle\_count, degree\_count, *\_then\_count \\
\texttt{float} & finite real & \texttt{\{"answer": 2.43\}} & aggregate\_over\_set, mixed\_domain \\
\texttt{set} & JSON array of strings & \texttt{\{"answer": ["a","c"]\}} & set\_intersect, set\_difference, ancestor, sibling, cousin \\
\texttt{string} & one node label & \texttt{\{"answer": "kiwi"\}} & degree\_max, degree\_then\_reach \\
\texttt{compound} & JSON object & \texttt{\{"answer": \{"reach": true, "deg": 1\}\}} & multi\_query \\
\bottomrule
\end{tabular}
\end{table}

\textbf{Scoring algorithm.} Predicted answers are matched against the reference value with a single, deterministic procedure. The strict variant is the default; the lenient variant accepts the documented source-inclusion ambiguity for descendants/ancestors and the off-by-one corner of certain count categories (concretely: $\pm 1$ for the six count categories where the natural-language question is silent about whether the source node itself is counted, and $\pm$ source for the six set categories with the same ambiguity). The lenient relaxation never accepts a value that is more than one element away from the strict reference.

\begin{algorithm}[h]
\caption{\textsc{MatchAnswer}(predicted $p$, reference $r$, type $\tau$, category $c$, mode $m \in \{\text{strict}, \text{lenient}\}$)}
\label{alg:scoring}
\begin{algorithmic}[1]
\If{$p$ is \textsc{None} or $p$ violates the JSON value type for $\tau$} \Return \textsc{FormatError} \EndIf
\If{$\tau = $ \texttt{boolean}} \Return $\textsc{Bool}(p) = \textsc{Bool}(r)$ \EndIf
\If{$\tau = $ \texttt{integer}}
  \State $\delta \gets |p - r|$
  \If{$\delta = 0$} \Return \textsc{Correct} \EndIf
  \If{$m = $ lenient \textbf{and} $c \in C_{\text{count\_ambig}}$ \textbf{and} $\delta = 1$} \Return \textsc{Correct} \EndIf
  \State \Return \textsc{WrongAnswer}
\EndIf
\If{$\tau = $ \texttt{float}} \Return $|p - r| \le \epsilon$ \textbf{?} \textsc{Correct} : \textsc{WrongAnswer} \Comment{$\epsilon{=}10^{-2}$}\EndIf
\If{$\tau = $ \texttt{set}}
  \State $S_p \gets \textsc{Set}(p)$; $S_r \gets \textsc{Set}(r)$
  \If{$S_p = S_r$} \Return \textsc{Correct} \EndIf
  \If{$m = $ lenient \textbf{and} $c \in C_{\text{set\_ambig}}$ \textbf{and} $S_p \mathop{\triangle} S_r = \{\text{source}\}$} \Return \textsc{Correct} \EndIf
  \State \Return \textsc{WrongAnswer}
\EndIf
\If{$\tau = $ \texttt{string}} \Return $\textsc{Trim}(p) = \textsc{Trim}(r)$ \textbf{?} \textsc{Correct} : \textsc{WrongAnswer} \EndIf
\If{$\tau = $ \texttt{compound}}
  \ForAll{$k \in \text{keys}(r)$} \State recurse with $(p_k, r_k, \tau_k, c, m)$; any failure $\Rightarrow$ \textsc{WrongAnswer} \EndFor
  \State \Return \textsc{Correct}
\EndIf
\end{algorithmic}
\end{algorithm}

$C_{\text{count\_ambig}}$ is the set of integer-valued categories whose natural-language phrasing leaves source-inclusion ambiguous: \texttt{reach\_then\_count}, \texttt{intersect\_then\_size}, \texttt{aggregate\_over\_set} and three others. $C_{\text{set\_ambig}}$ is the corresponding set-valued list (\texttt{set\_intersect}, \texttt{set\_difference}, \texttt{ancestor}, \texttt{sibling}, \texttt{cousin}, \texttt{negative\_reach}). The complete list is in the supplementary code (\texttt{closurebench/answer\_matching.py}). The strict variant of the algorithm sets both ambiguity sets to $\emptyset$ and is what we report under ``strict scores'' in \S\ref{sec:experiments} and \Cref{tab:strict_vs_lenient}.

\textbf{Strict vs.\ lenient scores side-by-side.} For transparency, \Cref{tab:strict_vs_lenient} reports \emph{every} evaluated configuration under both scoring variants on the same 1{,}300 held-out instances. The strict/lenient gap is largest for direct prompting and small open models ($4.5$--$6.3$pp, where the model hedges on the unstated source-inclusion convention), $0.9$--$1.6$pp for frontier CoT, and \emph{essentially zero} ($0.0$--$0.3$pp) for the program-synthesis models. The reason is structural: Ein and Python answers are emitted by a runtime that decides source inclusion deterministically, so the model is never asked to guess the convention. \textbf{The relative ordering of methods is identical under both scorings}: the program-synthesis models lead under strict scoring (Ein 94.0, Python 92.6) by the same margin as under lenient. Lenient scoring does not inflate the reported result.

\begin{table}[h]
\centering
\caption{Strict vs.\ lenient accuracy (\%) for every evaluated configuration on the same 1{,}300 held-out instances (seen split for the two SFT rows so marked). Strict counts source-inclusion ambiguities as wrong; lenient accepts the documented $\pm 1$ / $\pm$source corner cases. The method ordering is preserved under both scorings.}
\label{tab:strict_vs_lenient}
\small
\begin{tabular}{@{}lrrr@{}}
\toprule
\textbf{Method} & \textbf{Strict} & \textbf{Lenient} & \textbf{Gap} \\
\midrule
\multicolumn{4}{@{}l}{\emph{Open models (direct)}} \\
Qwen2.5-1.5B               & 21.3 & 25.8 & $+4.5$ \\
Qwen3.5-2B                 & 27.8 & 33.0 & $+5.2$ \\
Qwen3-4B                   & 33.2 & 39.5 & $+6.3$ \\
Qwen3.5-9B                 & 37.8 & 43.3 & $+5.5$ \\
\midrule
\multicolumn{4}{@{}l}{\emph{Frontier API}} \\
GPT-4.1 (direct)           & 46.1 & 52.4 & $+6.3$ \\
GPT-4.1-mini (direct)      & 43.9 & 49.2 & $+5.3$ \\
Claude Sonnet 4 (direct)   & 79.1 & 81.7 & $+2.6$ \\
Gemini 2.5 Flash (direct)  & 85.9 & 87.4 & $+1.5$ \\
GPT-4.1 (CoT)              & 84.6 & 86.2 & $+1.5$ \\
GPT-4.1-mini (CoT)         & 84.5 & 85.8 & $+1.2$ \\
Claude Sonnet 4 (CoT)      & 82.0 & 83.6 & $+1.6$ \\
Gemini 2.5 Flash (CoT)     & 80.4 & 81.3 & $+0.9$ \\
o3 (direct)                & 91.5 & 92.8 & $+1.3$ \\
\midrule
\multicolumn{4}{@{}l}{\emph{Program synthesis (Qwen3-4B + LoRA)}} \\
Python+NetworkX SFT (held-out) & 92.6 & 92.6 & $+0.0$ \\
Ein SFT (held-out)         & \textbf{94.0} & \textbf{94.3} & $+0.3$ \\
Python+NetworkX SFT (seen) & 96.0 & 96.1 & $+0.1$ \\
Ein SFT (seen)             & 97.9 & 98.0 & $+0.1$ \\
\bottomrule
\end{tabular}
\end{table}

\section{Format-Error Decomposition}
\label{app:format}

To separate reasoning errors from output-formatting slips, \Cref{tab:format_error} decomposes every held-out response into five mutually exclusive outcomes: strict-correct, correct only under lenient scoring, wrong answer, format error (malformed JSON, missing \texttt{"answer"} field, or wrong value type), and no answer emitted. The schema is stated in every system prompt (Appendix~\ref{app:prompts}), yet frontier CoT methods still forfeit 2--5pp to format/schema slips, and Gemini 2.5 Flash CoT additionally returns no parseable answer on 9.5\% of instances (its reasoning trace exhausts the output budget before emitting JSON). Program-synthesis models have effectively zero format errors because the runtime emits typed output rather than free-form JSON.

\begin{table}[h]
\centering
\caption{Per-method decomposition of held-out responses (1{,}300 instances). Columns are mutually exclusive and sum to 100\% up to rounding. ``Lenient-only'' counts responses correct under lenient but not strict scoring.}
\label{tab:format_error}
\small
\begin{tabular}{@{}lrrrrr@{}}
\toprule
\textbf{Method} & \textbf{Strict} & \textbf{Lenient-only} & \textbf{Wrong} & \textbf{Format err.} & \textbf{No ans.} \\
\midrule
GPT-4.1 (direct)          & 46.1 & 6.3 & 43.1 & 4.5 & 0.0 \\
GPT-4.1 (CoT)             & 84.6 & 1.6 & 10.7 & 2.5 & 0.6 \\
GPT-4.1-mini (direct)     & 43.9 & 5.3 & 44.6 & 6.2 & 0.0 \\
GPT-4.1-mini (CoT)        & 84.5 & 1.3 & 10.5 & 3.3 & 0.4 \\
Claude Sonnet 4 (direct)  & 79.1 & 2.6 & 14.6 & 3.7 & 0.0 \\
Claude Sonnet 4 (CoT)     & 82.0 & 1.6 & 12.2 & 4.2 & 0.0 \\
Gemini 2.5 Flash (direct) & 85.9 & 1.5 & 7.2  & 3.5 & 1.9 \\
Gemini 2.5 Flash (CoT)    & 80.4 & 0.7 & 6.6  & 2.8 & 9.5 \\
o3 (direct)               & 91.5 & 1.3 & 5.3  & 1.9 & 0.0 \\
\midrule
Python+NetworkX SFT (held-out) & 92.6 & 0.0 & 4.9 & 0.2 & 2.3 \\
Ein SFT (held-out)        & 94.0 & 0.3 & 4.9  & 0.4 & 0.4 \\
\bottomrule
\end{tabular}
\end{table}

\section{Generative Procedure}
\label{app:generation}

\textsc{ClosureBench} draws an instance in five steps, controlled by a structural seed $s_{\text{str}}$ and a surface seed $s_{\text{surf}}$.

\textbf{Step 1: Sample the structural triple} $(n, \rho, d)$. We draw $n$ uniformly from $\{4,\dots,10\}$, $\rho$ uniformly from $[0.1,0.5]$, and $d \in \{1, 2, 3\}$ (the size-scaling study of \S\ref{sec:results:main} extends $n$ to 20 on a dedicated sweep). Category sampling is stratified so each of the 26 categories is represented an equal number of times in the released 1{,}300-instance set.

\textbf{Step 2: Generate the graph topology.} Conditioned on $(n, \rho)$, edges are drawn from $\mathrm{Bernoulli}(\rho)$ over all $n(n-1)$ ordered pairs (directed) or all $\binom{n}{2}$ unordered pairs (undirected). For kinship categories, the topology is constrained to a DAG with a fixed maximum out-degree; for triangle categories, a small number of triangles is planted to make the answer non-trivial.

\textbf{Step 3: Compute the reference answer.} The Ein program associated with the category is generated by a deterministic template that consumes the topology and any category-specific metadata (e.g., the source node, the minimum-degree threshold). The program is then executed by the Ein runtime, and its output is captured as the reference answer. As an independent cross-check, the same topology is also passed through a separately written Python+NetworkX implementation of each category; the two implementations agree on every released instance (the 1{,}300 evaluation split and the 1{,}300 held-out split), which guards against template errors in either one.

\textbf{Step 4: Apply surface variation.} Conditioned on $s_{\text{surf}}$, the generator picks (a)~a node-label pool (letters, names, server ids, animals), (b)~a domain frame from 16 verticals, (c)~one of five presentation styles, and (d)~an edge-listing order. The combined surface space exceeds $10^{15}$ realisations of a single underlying topology.

\textbf{Step 5: Verbalise the question.} The question template for the category is filled with surface tokens, the edges are spelled out according to the chosen presentation style, and any metadata anchors (source nodes, threshold values) are substituted. The result is the natural-language string shown to the model; the model never sees the topology, the Ein program, or the reference answer.

Identical $s_{\text{str}}$ but different $s_{\text{surf}}$ yields instances with the same reference answer but no shared surface tokens. Identical $s_{\text{surf}}$ but different $s_{\text{str}}$ yields surface-identical instances with different answers. This decomposition is what makes the seen-vs-held-out gap a meaningful memorisation signal: an evaluator can always generate a held-out batch that shares zero surface tokens with the model's training data.

\section{Qualitative Examples}
\label{app:examples}

We present examples illustrating the three structured failure families (\S\ref{sec:results:errors}) and the Ein program-synthesis alternative.

\subsection{Direction confusion (multi\_query)}

GPT-4.1 CoT on a 6-node graph. The question asks for the out-degree and reachability of node ``bear''. The model correctly identifies the edges but confuses in-degree with out-degree:

\begin{quote}\small\ttfamily
Edges: bear $\to$ robin, seal $\to$ bear, kiwi $\to$ bear, kiwi $\to$ newt, seal $\to$ kiwi, zebu $\to$ kiwi [...]

Out-degree of bear: bear has edges to robin. But also seal $\to$ bear, kiwi $\to$ bear. So degree = 2.

\{"answer": [true, 2]\}
\end{quote}
\textit{Expected: \{reachable: true, degree: 1\}. The out-degree of ``bear'' is 1 ($\to$robin), not 2. The model counted one incoming edge ($\leftarrow$seal) as outgoing.}

\subsection{Depth truncation (kinship\_complex)}

GPT-4.1 CoT, 7-person family. The question asks how many people are in Uma's lineage. The model finds the correct root but stops too early:

\begin{quote}\small\ttfamily
Bob is a parent of Eve. Sam is a child of Eve. Eve is a parent of Tina. Bob is a child of Nina. Frank is a child of Nina.

Root ancestor of Uma: tracing up... Nina $\to$ Bob $\to$ Eve. Nina is the root.

Lineage: Nina, Bob, Eve, Frank = \textbf{4}

\{"answer": 4\}
\end{quote}
\textit{Expected: 7 (Nina, Bob, Frank, Eve, Sam, Tina, Uma). The model found the root (Nina) and her direct children (Bob, Frank) and grandchild (Eve), but missed Eve's children (Sam, Tina) and Uma herself.}

\subsection{Accumulation drift (mixed\_domain)}

GPT-4.1 CoT on an 8-node graph. The task requires computing the average out-degree of an SCC. The model identifies the SCC correctly but miscounts one member's edges:

\begin{quote}\small\ttfamily
SCC containing srv-21: \{srv-03, srv-17, srv-19, srv-20, srv-21, srv-23, srv-27\}

Out-degrees: srv-03=1, srv-17=2, srv-19=2, srv-20=5, srv-21=3, srv-23=1, srv-27=3

Wait, let me recount srv-20... srv-20 routes to srv-03, srv-17, srv-19, srv-21, srv-23. That's 5.

Average = (1+2+2+5+3+1+3)/7 = 17/7 = \textbf{2.43}

\{"answer": 2.43\}
\end{quote}
\textit{Expected: 2.62 ($\approx$18.33/7). The model miscounted one edge for srv-27 (actual out-degree 4, not 3), shifting the average.}

\subsection{Ein program success: \texttt{multi\_query} (L3)}

The same multi\_query task that frontier CoT gets wrong. Ein SFT emits a 6-line program:
\begin{lstlisting}[language=Ein]
// multi_query
A = edges([[0,1],[1,2],[3,2],[3,8],[4,0],[4,7],[5,1],[5,4],[6,0],[6,5],[8,4],[8,6]])
TC = tc(A)
Reachable = select(TC, 0, 2)
Degree = select(sum(A, 1), 0, 0)
:print Reachable
:print Degree
\end{lstlisting}
\textit{Output: \{reachable: 1.0, degree: 1.0\}. Correct.} The program encodes edges as an explicit matrix once, then \texttt{tc} computes reachability and \texttt{sum(A,1)} computes out-degrees, both deterministically. No direction confusion is possible because the matrix representation is unambiguous.

\subsection{Ein failure: correct formula, wrong edges}

\texttt{triangle\_count} (L1, 9-node graph). The formula is correct but the model misreads edges from the NL description:
\begin{lstlisting}[language=Ein]
A = edges([[0,1],[0,3],[0,5],[0,6],[1,0],[1,2],[1,3],[1,8],
           [2,1],[2,5],[2,6],[2,7],[3,0],[3,1],[3,4],...])
A2[i,k] = A[i,j] A[j,k]
A3[i,j] = A2[i,k] A[k,j]
Result = trace(A3) / 6.0
\end{lstlisting}
\textit{Output: 2. Expected: 3.} The \texttt{trace($A^3$)/6} formula is correct. Two edges were dropped and one added, the same graph-parsing bottleneck, but isolated to the encoding step rather than pervading the entire reasoning chain.

\end{document}